\documentclass[10pt, a4paper]{article}
\usepackage{lrec2022} % this is the new LREC2022 Style
\usepackage{multibib}
\newcites{languageresource}{Language Resources}
\usepackage{graphicx}
\usepackage{tabularx}
\usepackage{soul}
\usepackage{titlesec}
\titleformat{\section}{\normalfont\large\bfseries\center}{\thesection.}{1em}{}
\titleformat{\subsection}{\normalfont\SmallTitleFont\bfseries\raggedright}{\thesubsection.}{1em}{}
\titleformat{\subsubsection}{\normalfont\normalsize\bfseries\raggedright}{\thesubsubsection.}{1em}{}
\renewcommand\thesection{\arabic{section}}
\renewcommand\thesubsection{\thesection.\arabic{subsection}}
\renewcommand\thesubsubsection{\thesubsection.\arabic{subsubsection}}
\usepackage{epstopdf}
\usepackage[utf8]{inputenc}
\usepackage{hyperref}
\usepackage{xstring}
\usepackage{kotex}
\usepackage{color}

\title{Korean-Specific Dataset for Table Question Answering}

\name{Changwook Jun, Jooyoung Choi, Myoseop Sim, Hyun Kim, Hansol Jang, Kyungkoo Min} 

\address{LG AI Research \\
         ISC, 30, Magokjungang 10-ro, Gangseo-gu, Seoul, 07796, Korea \\
         \{cwjun, jooyoung.choi, myoseop.sim, hyun101.kim, hansol.jang, mingk24\}@lgresearch.ai\\}

\abstract{
Existing question answering systems mainly focus on dealing with text data. However, much of the data produced daily is stored in the form of tables that can be found in documents and relational databases, or on the web. To solve the task of question answering over tables, there exist many datasets for table question answering written in English, but few Korean datasets. In this paper, we demonstrate how we construct Korean-specific datasets for table question answering: Korean tabular dataset is a collection of 1.4M tables with corresponding descriptions for unsupervised pre-training language models. Korean table question answering corpus consists of 70k pairs of questions and answers created by crowd-sourced workers. Subsequently, we then build a pre-trained language model based on Transformer and fine-tune the model for table question answering with these datasets. We then report the evaluation results of our model. We make our datasets publicly available via our GitHub repository and hope that those datasets will help further studies for question answering over tables, and for the transformation of table formats.
 \\ \newline  }

\begin{document}

\maketitleabstract

\section{Introduction}

The task of question answering is to correctly answer to given questions, which requires a high level of language understanding and machine reading comprehension abilities. As pre-trained language models on Transformer~\cite{vaswani2017attention} have brought significant improvements in performance in many natural language processing tasks, there have been many studies in machine reading comprehension (MRC) and question answering (QA) tasks~\cite{devlin2018bert,lan2019albert,yang2019xlnet,yamada2020luke,liu2019roberta,clark2020electra}. There are Stanford Question Answering Dataset (SQuAD) benchmarks~\cite{rajpurkar2016squad,rajpurkar2018know}, well-known machine reading comprehension benchmarks in the NLP area, that involve reasoning correct answer spans in the evidence document to a given question. Since SQuAD datasets are composed of pairs of natural language questions and answers, and the corresponding documents are unstructured textual data, the task is mainly to focus on predicting answers from plain texts.

Much of the world’s information produced daily is stored in structured formats such as tables in databases and documents, or tables on the web. Question answering over these structured tables has been generally considered a semantic parsing task in which a natural language question is translated to a logical form that can be executed to generate the correct answer~\cite{pasupat2015compositional,zhongSeq2SQL2017,dasigi2019iterative,wang2019rat,rubin2020smbop}. There have been many efforts to build semantic parsers with supervised training datasets such as WikiSQL~\cite{zhongSeq2SQL2017} consisted of pairs of questions and structured query language (SQL) and Spider dataset~\cite{yu2018spider} that aims for a task of converting text to SQL. However, it is expensive to create such data, and there are challenges in generating logical forms. In recent years, a few studies attempt the task of question answering over tables without generating logical forms~\cite{herzig2020tapas,yin2020tabert,chen2020hybridqa,zayats2021representations,zayats2021representations}. They introduce approaches of pre-trained language models based on BERT~\cite{devlin2018bert} to learn representations of natural language sentences and structured tables jointly by extending embeddings, and these models achieve strong performance on the semantic parsing datasets.

In this paper, for the Korean-specific table question answering task, we present KO-TaBERT, a new approach to train BERT-based models that learn jointly textual and structured tabular data by converting table structures. To address this, we first create two datasets written in the Korean language: the tabular dataset contains conversion formats of around 1.4M tables extracted from Korean Wikipedia documents for pre-training language models, and the table question answering dataset for fine-tuning the models. The table question answering dataset consists of 70k pairs of questions and answers, and the questions are generated by crowdsourced workers considering question difficulty. Additionally, we introduce how structured tables are converted into sentence formats. The conversion formats play a crucial role for models to learn table structural information effectively without changing embeddings. Second, we follow BERT architecture~\cite{devlin2018bert} to pre-train a language model with the converted strings from millions of tables, and fine-tune models on the table question answering dataset. All resources we create in this study are released via our GitHub repository\footnote{https://github.com/LG-NLP/KorWikiTableQuestions}.

We evaluate our model on the downstream task of table question answering. For performance evaluation, we create the test dataset that includes around 10k question-answer pairs related to tables in Korean Question Answering Dataset (KorQuAD 2.0)\footnote{https://korquad.github.io/}, and 20\% splits of the crowdsourced dataset. KO-TaBERT achieves EM 82.8\% and F1 86.5\% overall. Comparisons of the model performance according to question difficulty and different table conversion formats are also reported in this study.

We summarize our main contributions as follows:
\begin{itemize}
\item We construct two Korean-specific datasets: Korean Wikipedia Tabular dataset (KorWikiTabular) consisting of 1.4M tables that are converted into sentence strings containing tabular structural information, and Korean Wikipedia Table Questions dataset (KorWikiTQ) including 70k pairs of questions and answers generated according to question difficulty by paid crowdsourced workers.
\item We introduce approaches to converting tabular data into sentence strings, which allows models to represent table structural properties and information.
\item We present KO-TaBERT, a pre-trained language model that learns syntactic and lexical information and as well structural information from tables.
\item We build table question answering models based on the pre-trained language model using our datasets. Our model is evaluated on the table-related subset of the KorQuAD 2.0 corpus, and they can be treated as baselines for future research.
\end{itemize}

\section{Related Works}
There have been many studies to reason the correct answer to a given question over tables. A semantic parsing task is applied for question answering over tables, which translates natural language questions to logical forms such as SQL. Several works~\cite{hwang2019comprehensive,lyu2020hybrid,lin2020bridging,cao2021lgesql} introduce approaches that leverage BERT~\cite{devlin2018bert}, a pre-trained language model (PLM), in the text-to-SQL task, since PLMs with a deep contextualized word representation have led noticeable improvements in many NLP challenges. Word contextualization has contributed to improving accuracy for the generation of logical forms, but there still remains difficulties to generate logical forms obeying decoding constraints. There is also a limitation that the text-to-SQL approaches require table data to be stored in a database.
Many recent studies have shown that supervised question answering models successfully reason over tables without generating SQL. TAPAS~\cite{herzig2020tapas} is a new approach to pre-training a language model that extends BERT by inserting additional embeddings to better understand tabular structure. Similarly, TURL~\cite{deng2020turl} implements structure-aware Transformer encoder to learn deep contextualised representations for relational table understanding. Masked Entity Recovery as a novel pre-training objective is proposed to capture complex semantics knowledge about entities in relational tables. TABERT~\cite{yin2020tabert} and TABFACT~\cite{chen2019tabfact} also use BERT architecture to jointly understand contextualised representations for textual and tabular data. Unlike other approaches, TABERT encodes a subset of table content that is most relevant to a question in order to deal with large tables.
In this study, we propose a BERT-based approach to learn deep contextualised representations of table structural information via conversion of table formats for the task of table question answering.

\section{Korean Table Question Answering Dataset}
In this study, we pre-train a language model with structured data extracted from Korean Wikipedia for the task of question answering over table, since pre-training models have shown significant improvements in many natural language understanding tasks. For pre-training input data, we generate \textit{KorWikiTabular} containing pairs of structured tabular data and related texts to the tables. We also create \textit{KorWikiTQ}, a corpus of question-answer pairs with tables in which the answers are contained for the table question answering task.

\subsection{Tabular Dataset for Pre-training}
\label{pre-training_dataset}
\begin{figure*}[t]
\begin{center}
\includegraphics[width=12cm]{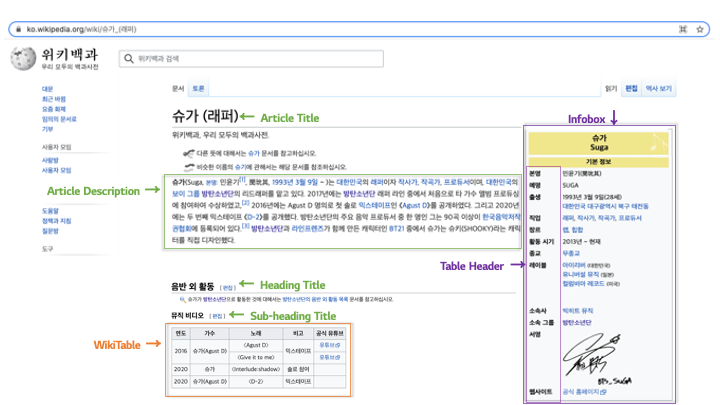}
\caption{Example of \textit{Infobox} and \textit{WikiTable} in a Wikipedia document}
\label{fig.1}
\end{center}
\end{figure*}

We collect about 1.4M tables (\textit{T}) from the Korean Wikipedia dump in order to pre-train a Transformer-based language model for tabular contextualised embeddings. Since we hypothesis that descriptions (\textit{D}) in the article of wikipedia that contains a table benefit building improved representations of the table for question answering, pairs of tables and their description texts are extracted from Wikipedia. As pre-training inputs for tabular contextualization, we extract \textit{Infobox} which is formatted as a table on the top right-hand corner of Wikipedia documents to summarise the information of an article on Wikipedia, and \textit{WikiTable} as shown in Figure \ref{fig.1}. As description texts for a table, an article title, the first paragraph in the article, table captions if exist, heading and sub-heading titles for the article are considered as the pre-training dataset. That is, \textit{D = \{d\textsubscript{1},d\textsubscript{2}, ..., d\textsubscript{n}}\}.

\begin{figure}[!h]
\begin{center}
\includegraphics[width=8cm]{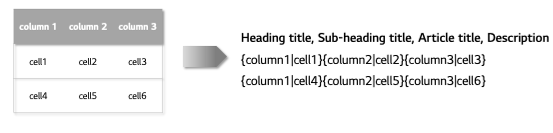}
\caption{Converting \textit{WikiTable} format into string sequences for pre-training input. The converted table strings are added with descriptions for the Wikipedia article.}
\label{fig.2}
\end{center}
\end{figure}

For pre-training a language model, we generate input sequences by converting the extracted infoboxes and wikitables into sentence string texts. Figure \ref{fig.2} explains how a table format is converted into sentence strings when the table is a relational table that contains columns (\textit{T\textsubscript{c}}) or fields describing rows (\textit{T\textsubscript{r}}) of data. Each table is formed in two-dimensional tabular data consisting of columns and rows, denoted as \textit{T = \{t\textsubscript{(c1,r1)},t\textsubscript{(c2,r1)},t\textsubscript{(cn,r1)} ..., t\textsubscript{(c1,rm)},t\textsubscript{(c2,rm)},t\textsubscript{(cn,rm)}}\} when \textit{n} and \textit{m} are sizes of column and row respectively. Table column headers are identified with rows and cells, and then the columns are concatenated with corresponding cells with predefined special characters, which allows the model to learn structural relations between cells of columns and table headers.

For \textit{Infobox}, the table headers mainly located in the first column are also recognised using the $<$th$>$ tags, then they are converted into the string sequences associated with relevant cells like wikitables.

\subsection{Crowdsourcing for Table Question Answering Corpus}
\label{fine-tuning_dataset}
\begin{figure*}[t]
\begin{center}
\includegraphics[height=7cm]{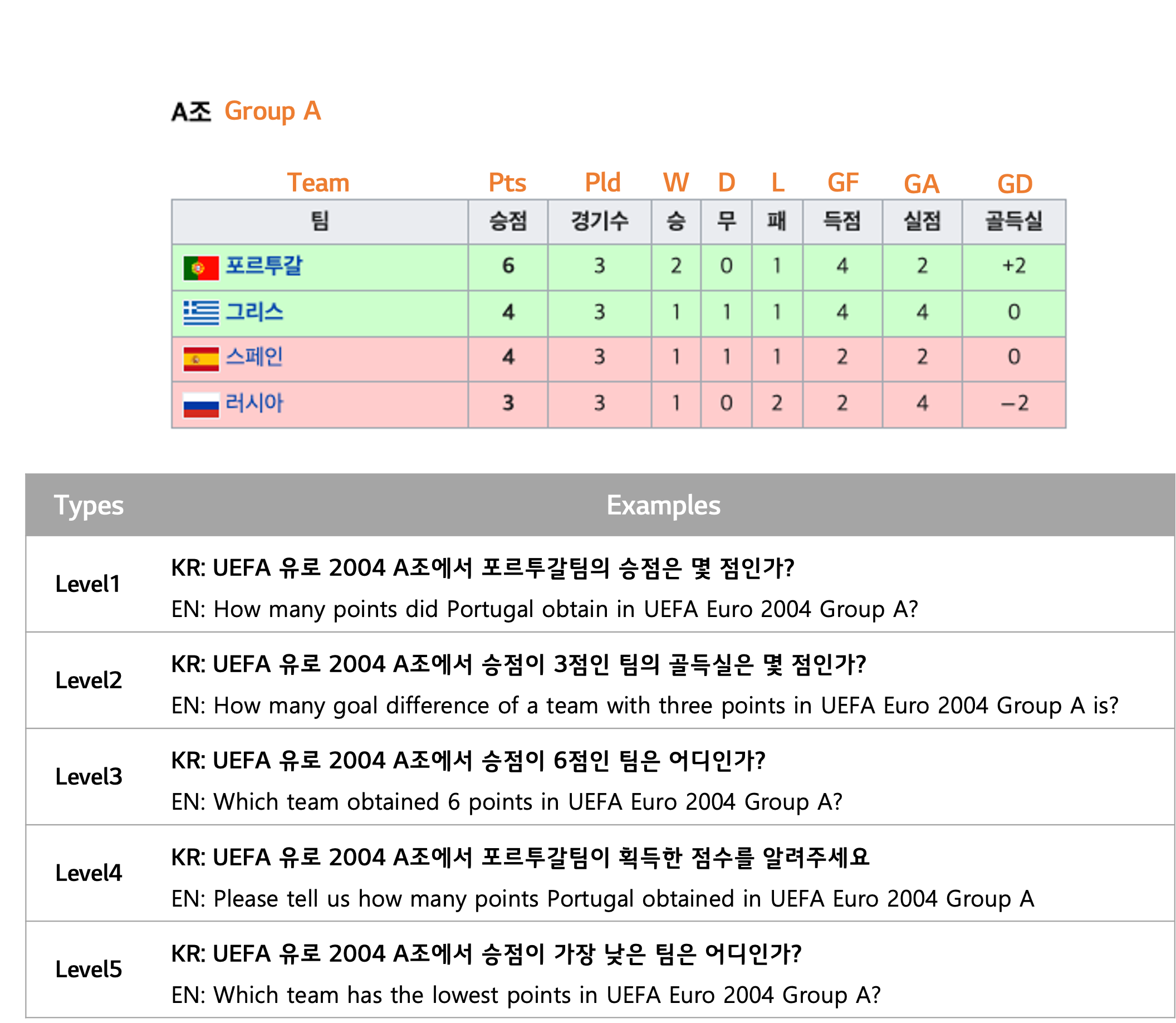}
\caption{Examples of a natural language question set related to a table for football match-results in UEFA Euro 2004 according to question levels}
\label{fig.3}
\end{center}
\end{figure*}

Unfortunately, there do not exist datasets written in Korean aiming for the table question-answering task publicly available. Even though KorQuAD 2.0 ~\cite{lim2019korquad1} contains some questions that their answers are related to tables, they are insufficient to train models for the task. 

For conducting the downstream task of question answering over tables, we create a large dataset consisting of 70k pairs of questions and answers by hiring paid, crowdsourced workers. We select about 20k tables that contain rows greater than 5 and less than 15 for this crowdsourcing task. Larger tables including columns greater than 10 are ignored, since the length of the converted sentence strings from a table may exceed the maximum sequence length of 512 tokens. This is because column headers are repeatedly inserted into the strings as the number of table headers increases. 

We define the five-question variations by extending to the previous work~\cite{park2020korean}, considering question difficulties that our model can predict answers as follows:

\begin{itemize}
    \item Level1: Question [column$-$others] where [column$-$base] has [value] 
    \item Level2: Question [column$-$others] where [column$-$base] has [condition]
    \item Level3: Question [column$-$base] where [column$-$others] has [value]
    \item Level4: Variation of the questions in other levels
    \item Level5: Question min or max in [column$-$base] where [column$-$others] has [value of numbers, dates, ranks, etc.]
\end{itemize}

Figure \ref{fig.3} describes question types generated regarding the question difficulties. Our crowd-sourced workers generate the five types of natural language questions for a given table. As shown in Figure \ref{fig.3}, an example question of Level1 is KR: UEFA 유로 2004 A조에서 포르투갈팀의 승점은 몇 점인가? (EN: \textit{how many points did Portugal obtain in UEFA Euro 2004 Group A?}): querying values of other columns such as \textit{Points (Pts)} when the value of the base column \textit{Team} is Portugal. In this case, we intend an answer value to be located on the right of the base column. For Level3, on the contrary, an answer is located in the left column from the base column as shown in the example, KR: UEFA 유로 2004 A조에서 승점이 6점인 팀은 어디인가?
 (EN: \textit{which team obtained 6 points in UEFA Euro 2004 Group A?}). We include these various types of questions for improvement of model performance since our model tends to produce a lower performance on the type of Level3 question than Level1. A variety of sentence expressions such as honorific, informal, colloquial, and written formats are used in the Korean language. To enhance the robustness of our model to various types of natural language questions, we incorporate sentence variations of the questions in the other levels by changing adverbs or prepositions, replacing them with synonyms, inverting word orders, or converting sentences into honorific expressions, formal and informal languages. Numeric data such as numbers, ranks, money, or dates are stored in tables. For these tables, we generate questions related to operations (minimum, maximum, count, or average) for columns, for example, KR: 승점이 가장 낮은팀은 어디? (EN: \textit{Which team has \underline{the lowest points} in UEFA Euro 2004 Group A?}).

 The Korean table question-answering dataset consists of table contexts (\textit{C}) in the form of two-dimensional lists, natural language questions (\textit{Q}), and answers (\textit{A}), along with URLs (\textit{U}) and titles (\textit{T}) of Wikipedia documents where tables are contained. This dataset is provided in the JSON format.

\section{Modeling of Table Question Answering}
In this study, we pre-train a language model with the converted tabular dataset described in Section \ref{pre-training_dataset} for a better understanding of structural information and as well syntactic and lexical information considering the downstream task of table question-answering. We also fine-tune the model using the table question answering corpus. We evaluate our model, and the fine-tuning results in the task of question-answering over tables are described in detail.

\subsection{Pre-training Language Model}
We use the Transformer~\cite{vaswani2017attention} approach for pre-training a language model. In specific, we follow the original architecture of BERT~\cite{devlin2018bert} which is a transformer-based pre-trained language model, which learns context in text using masked language modeling (MLM) and the next sentence prediction (NSP) objectives for self-supervised pre-training. BERT\textsubscript{Base} model is adopted in our experiments, and we use the same percentage of 15\% for masking table cells or text segments for the MLM training, but the NSP objective is not used. We build a new vocabulary of 119,547 Korean word-pieces.
We generate input sequences with the special tokens [CLS] and [SEP], and description texts including relevant sentences such as the article title and heading titles are inserted between them. Tabular data converted to string sequences is followed by the [SEP] token. During a pre-training procedure, our model learns contextual representations jointly from unstructured natural language sentences and structured tabular data. Details of hyperparameters that we used for pre-training KO-TaBERT are summarised in Appendix~\ref{sec:appendix_1}.

\subsection{Fine-tuning Model}
For fine-tuning of the table question-answering task, we prepare datasets from different sources: the table question-answering dataset created by crowd-sourced workers described in ~\ref{fine-tuning_dataset}, and about 2k pairs of questions and answers related to tables selected from KorQuAD 2.0 corpus. We split the selected KorQuAD 2.0 corpus and the crowdsourced dataset with a 20\% ratio respectively as the test dataset for evaluation. The rest of the data is used for fine-tuning training.  
We train a question-answering model on the dataset. Similar to SQuAD which is a major extractive question-answering benchmark~\cite{rajpurkar2016squad,rajpurkar2018know}, the fine-tuning model aims to predict the correct span of an answer defined as a start and end boundaries for a given question. We describe the experimental details for the downstream task in Appendix~\ref{sec:appendix_2}.

\subsection{Evaluation and Results}
\label{eval_results}
We evaluate our model on the test dataset consisting of the subsets from the KorQuAD 2.0 and the crowdsourced table question-answering datasets, which are not used during the training of the model. As for evaluation metrics, Exact Match (EM) scores if each character of a predicted answer is exactly the same as the ground truth and F1-score (F1) that calculates token overlaps between a predicted answer and the ground-truth as F1 are used in our experiments.

\begin{table}[h]
\centering
\begin{tabular}{l cc cc }
\hline Dataset source && EM && F1 \\
\hline
KorQuAD 2.0 && 64.5 && 74.5 \\
Crowd-sourced  && 83.9 && 87.2 \\
\hline
Overall && 82.8 && 86.5 \\
\hline
\end{tabular}
\caption{\label{performance_total} Performance with EM and F1 scores according to the data sources of the test dataset for the task of table question answering.}
\end{table}

The performance of the table question-answering model is summarised in Table~\ref{performance_total}. Our model achieves EM 82.8\% and F1 86.5\% overall. We compare the performance of different data sources in the test dataset. We observe that our model performs better in the crowdsourced dataset than in the KorQuAD 2.0 corpus. With detailed error analysis, we find that there are a few parsing errors and table format conversion errors.

\begin{table}[h]
\centering
\begin{tabular}{l cc cc }
\hline Question difficulty && EM && F1 \\
\hline
Level1 && 89.6 && 93.1 \\
Level2 && 89.1 && 92.3 \\
Level3 && 86.1 && 89.4 \\
Level4 && 81.7 && 85.5 \\
Level5 && 67.8 && 70.6 \\
\hline
Overall && 83.9 && 87.2 \\
\hline
\end{tabular}
\caption{\label{performance_levels} Comparison of model performance according to each level of questions in the crowdsourced dataset.}
\end{table}

The test dataset consists of questions with their difficulties generated by crowdsourced workers as described in Section~\ref{fine-tuning_dataset}. The performance according to the levels of questions is described in Table~\ref{performance_levels}. The results show that the performance decreases as the difficulty of questions increases. In particular, the table question-answering model performs poorly on the Level5 questions that require comprehensive machine understanding capabilities such as ordering and counting table cell values, and operations such as minimum, maximum or average in a table column. 

\begin{figure}[!h]
\begin{center}
\includegraphics[width=8cm]{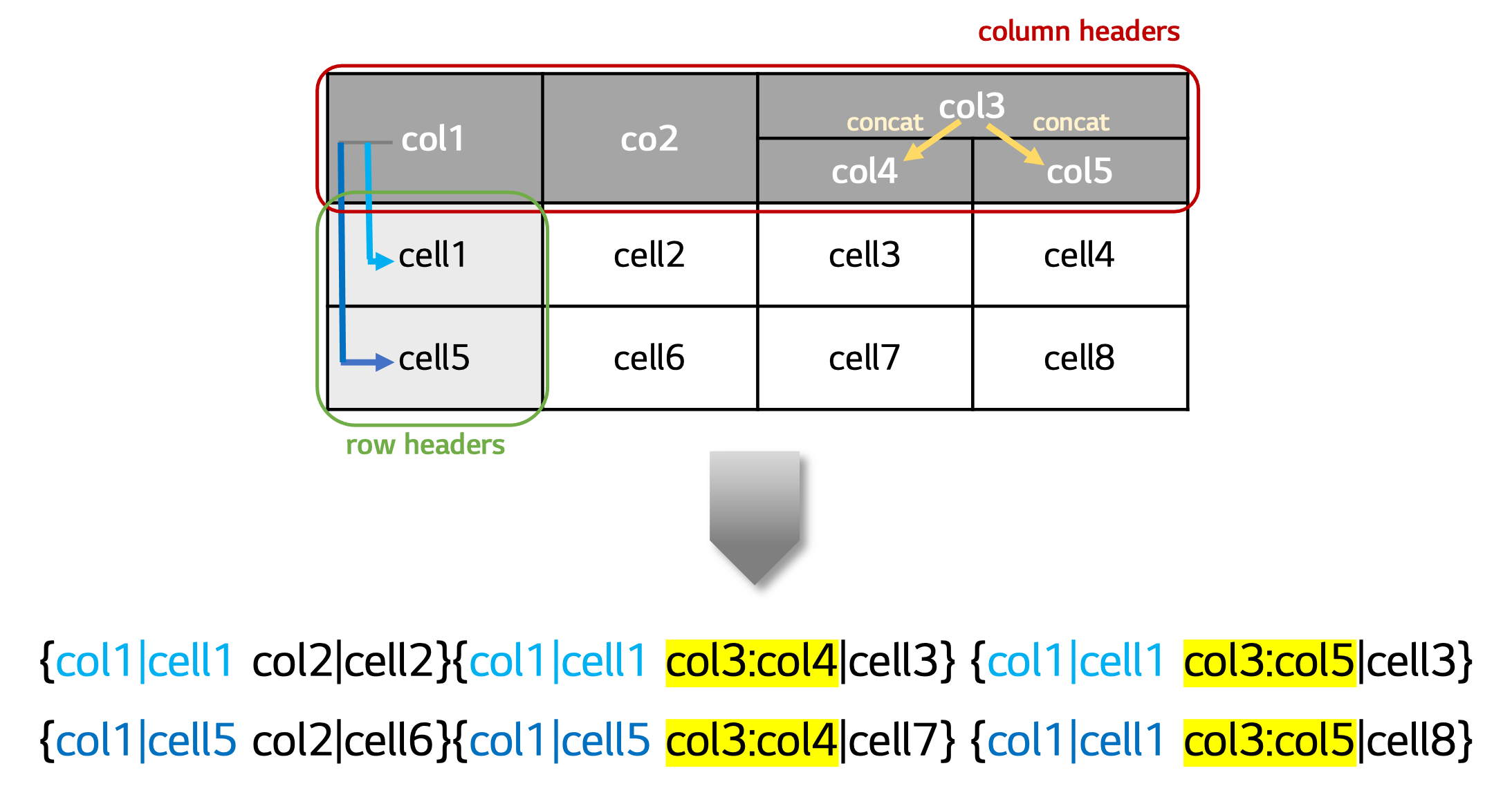}
\caption{Example of the new conversion approach for complicated structured tables that consisting of merged and multi column headers into sentence strings.}
\label{fig.4}
\end{center}
\end{figure}

We conduct additional experiments with the new approach of table format conversion by mapping the cell values in the first column (called row header) into the first column header. As illustrated in Figure~\ref{fig.4}, column 1 accompanied by the value of cell 1 is added to table cell values in the same row. This aims to apply important information in a table to the converted sentence strings by considering table structural characteristics. We also take account of tables informed of complicated structures. For converting those tables into sentence strings, we classify tables into single, multi, and merged column headings using $<$th$>$ tags. Then, column headers are concatenated if table columns have multi headers as shown in the example of Figure~\ref{fig.4}. In this conversion approach, the length of converted sentence strings increases since the headers are repeatedly inserted. Thus, we limit the word length of a converted sentence string from 250 to 300 tokens for pre-training input.

\begin{table}[h]
\centering
\begin{tabular}{l cc cc cc }
\hline Dataset source && Format && EM && F1 \\
\hline
KorQuAD 2.0 && v1 && 64.5 && 74.5 \\
KorQuAD 2.0 && v2 && \textbf{69.1} && \textbf{78.4} \\
\hline
\hline
Crowd-sourced && v1 && 83.9 && 87.2 \\
Crowd-sourced && v2 && \textbf{87.2} && \textbf{91.2} \\
\hline
\end{tabular}
\caption{\label{performance_modification} Comparison of model performance with different table parsing approaches. Format v1 is the table conversion described in Figure~\ref{fig.2}.}
\end{table}

We compare the performance of models pre-trained with different table conversion formats in Table~\ref{performance_modification}. Using the new format of table conversion considering structural complexity improves the performance of models on the table question answering task. The results indicate that the new table conversion format can effectively apply table structural features. 

\section{Conclusions}
In this paper, we introduce two new Korean-specific datasets, KorWikiTabular and KorWikiTQ for the task of table question-answering in the Korean language, and present KO-TaBERT, a pre-trained language model for the task. In particular, we demonstrate how tabular data is converted into linearlised texts containing structural information and properties. We construct a tabular dataset by extracting tables and converting them into sentence strings with tabular structural information for pre-training a language model. We also create a table question answering corpus with paid crowdsourced workers. The corpus consists of 70k pairs of questions and answers related to tables on Wikipedia articles, and those questions are generated specifically considering levels of question difficulty. We conduct experiments on the table question answering task. Our model achieves the best performance when converted table sentence strings include richly structural features. In future work, we aim to extend the model for complex question answering over texts and tables with the generation of multimodal questions to jointly handle question answering from textual and tabular sources. We hope that our datasets will help further studies for question answering over tables, and for the transformation of table formats.
\section{References}
\bibliographystyle{lrec2022-bib}
\bibliography{lrec2022-example}

% \section{Language Resource References}
% \label{lr:ref}
% \bibliographystylelanguageresource{lrec2022-bib}
% \bibliographylanguageresource{languageresource}

\appendix
\section*{Appendices}
\label{sec:appendix}
\addcontentsline{toc}{section}{Appendices}
\renewcommand{\thesubsection}{\Alph{subsection}}

\subsection{Pre-training Details}
\label{sec:appendix_1}
Table~\ref{table_hyperparameter} summarises hyperparameters that we use for pre-training KO-TaBERT. We use the maximum sequence length of 512, and the Adam optimization with the learning rate of 1e-4 is set for the model. Our model is trained on 64 TPU v3 for 1M steps with the batch size of 1024, and it takes about 5 days. 

\begin{table}[h]
\centering
\begin{tabular}{l c c }
\hline Hyper-parameters & KO-TaBERT \\
\hline
Number of layers & 12 \\
Hidden size & 768 \\
FFN inner hidden size & 3072 \\
Attention heads  & 12 \\
Attention head size & 64 \\
Dropout & 0.1 \\
Warmup steps & 10k \\
Learning rates & 1e-4 \\
Batch size & 1024 \\
Weight decay & 0.01 \\
Max steps  & 1M \\
Learning rate decay & Linear \\
Adam $\varepsilon$ & 1e-6 \\
Adam $\beta_{1}$ & 0.9 \\
Adam $\beta_{2}$  & 0.999 \\
\hline
Number of TPU & 64 \\
Training time  & 5 days \\
\hline
\end{tabular}
\caption{\label{table_hyperparameter} Hyperparameters for pre-training KO-TaBERT.}
\end{table}

\subsection{Fine-tuning Experiment}
\label{sec:appendix_2}
We describe the fine-tuning results of our model KO-TaBERT transferred to the task of question answering over tables in Section~\ref{eval_results}. For the experiments, we use the maximum sequence length of 512 and the Adam optimization with the learning rate of 5e-5. Our model is fine-tuned on 8 TPU v3 with a batch size of 32 and 3 epochs. Other hyperparameters are the same as the ones used for pre-training.

\end{document}